\documentclass[10pt,twocolumn,letterpaper]{article}
\pdfoutput=1

\usepackage{cvpr}
\usepackage{times}
\usepackage{epsfig}
\usepackage{graphicx}
\usepackage{amsmath}
\usepackage{amssymb}

\usepackage{algorithm}
\usepackage{algorithmic}
\usepackage{minibox}

\renewcommand{\algorithmicrequire}{\textbf{Input:}}
\renewcommand{\algorithmicensure}{\textbf{Output:}}
\newcommand{\argmax}{\operatornamewithlimits{argmax}}
\newcommand{\argmin}{\operatornamewithlimits{argmin}}

\usepackage[breaklinks=true,bookmarks=false]{hyperref}

\cvprfinalcopy 

\begin{document}

\title{Max-Margin Object Detection}

\author{Davis E. King\\
{\tt\small davis@dlib.net}
}

\maketitle

\begin{abstract}
Most object detection methods operate by applying a binary classifier to
sub-windows of an image, followed by a non-maximum suppression step where
detections on overlapping sub-windows are removed.  Since the number of
possible sub-windows in even moderately sized image datasets is extremely
large, the classifier is typically learned from only a subset of the
windows.  This avoids the computational difficulty
of dealing with the entire set of sub-windows, however, as we will show in this
paper, it leads to sub-optimal detector performance.  

In particular, the main contribution of this paper is the introduction of a new
method, Max-Margin Object Detection (MMOD), for learning to detect objects in
images.  This method does not perform any sub-sampling, but instead optimizes over all
sub-windows.  MMOD can be used to improve any object detection method which is
linear in the learned parameters, such as HOG or bag-of-visual-word models.
Using this approach we show substantial performance gains on three publicly
available datasets.  Strikingly, we show that a single rigid HOG filter can
outperform a state-of-the-art deformable part model on the Face Detection Data Set and Benchmark
when the HOG filter is learned via MMOD.
\end{abstract}


\section{Introduction}
\label{introduction}

Detecting the presence and position of objects in an image is a fundamental task in
computer vision.  For example, tracking humans in video or performing scene understanding 
on a still image requires the ability to reason about the number and position of 
objects.   While great progress has been made in recent years in terms of feature sets,  
the basic training procedure has remained the same.  
In this procedure, a set of positive and negative image windows are selected from
training images.  Then a binary classifier is trained on these windows.  
Lastly, the classifier is tested on images containing no targets of interest, and 
false alarm windows are identified and added into the training set.  The 
classifier is then retrained and, optionally, this process is iterated.

This approach does not make efficient use of the available training data since it trains on
only a subset of image windows.  Additionally,  windows partially overlapping an 
object are a common source of false alarms.  This training procedure makes it 
difficult to directly incorporate these examples into the training set since these
windows are neither fully a false alarm or a true detection.  Most importantly, 
the accuracy of the 
object detection system as a whole, is not optimized.  Instead, the accuracy of a binary
classifier on the subsampled training set is used as a proxy.  

In this work, we show how to address all of these issues.  In particular, we
will show how to design an optimizer that runs over all windows and optimizes
the performance of an object detection system in terms of the number of missed
detections and false alarms in the final system output.  Moreover, our
formulation leads to a convex
optimization and we provide an algorithm which finds the globally optimal 
set of parameters.  Finally, we test our method on three publicly available
datasets and show that it substantially improves the accuracy of the learned detectors.
Strikingly, we find that a single rigid HOG filter can outperform a state-of-the-art
deformable part model if the HOG filter is learned via MMOD.



\section{Related Work}

In their seminal work, Dalal and Triggs introduced the Histogram of Oriented Gradients (HOG)
feature for detecting pedestrians within a sliding window framework \cite{hog}.    
Subsequent object detection research has focused primarily on finding improved representations. 
Many recent approaches include features for part-based-modeling, methods for combining local features,
or dimensionality reduction \cite{huang, wu, schwartz, duan, tuzel}.
All these methods employ some form of binary classifier trained on positive and negative
image windows.  

In contrast, Blaschko and Lampert's research into structured output regression is the most similar 
to our own \cite{Blaschko08}.  As with our approach, they use a structural support 
vector machine formulation, which allows them to train on all window locations.
However, their training procedure assumes an image contains either
0 or 1 objects.  While in the present work, we show how to treat 
object detection in the general setting where an image may contain any number of objects.

\section{Problem Definition}

In what follows, we will use $r$ to denote a rectangular area of an image.  Additionally,
let $\mathcal{R}$ denote the set of all rectangular areas scanned by our object detection 
system.  To incorporate the common non-maximum suppression practice, we define a valid 
labeling of an image as a subset of $\mathcal{R}$ such that each
element of the labeling ``does not overlap'' with each other.   We use the following popular definition
of ``does not overlap'':  rectangles $r_1$ and $r_2$ do not
overlap if the ratio of their intersection area to total area covered is less than 0.5.  That is,
\begin{equation}
\label{eq:overlap}
\frac{Area(r_1 \cap r_2)}{Area(r_1 \cup r_2)} < 0.5.
\end{equation}
Finally, we use $\mathcal{Y}$ to denote the set of all valid labelings.  

Then, given an image $x$ and a window scoring function $f$, we can define the object detection
procedure as 
\begin{equation}
\label{eq:detector}
y^* = \argmax_{y \in \mathcal{Y}}\sum_{r \in y} f(x,r).
\end{equation}
That is, find the set of sliding window positions which have the largest scores
but simultaneously do not overlap.  This is typically accomplished with the
greedy peak sorting method shown in Algorithm~\ref{alg:detector}.
An ideal learning algorithm would find the window scoring function which 
jointly minimized the number of false alarms
and missed detections produced when used in Algorithm~\ref{alg:detector}.

\begin{figure}[ht]
\begin{center}
\centerline{\includegraphics[scale=0.35]{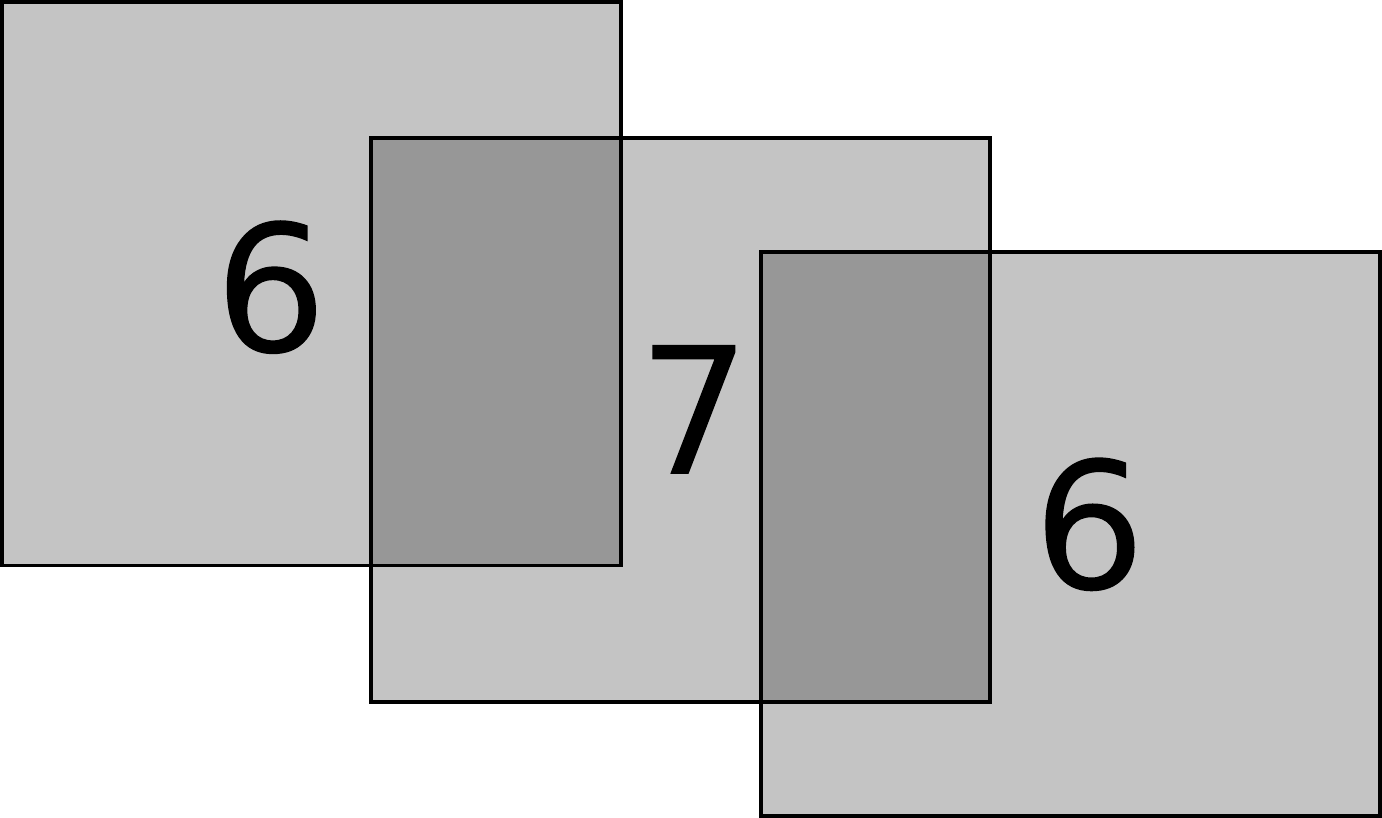}}
\caption{Three sliding windows and their $f$ scores. Assume non-max suppression rejects any rectangles which
touch.  Then the optimal detector would select the two outside rectangles, giving a total score of 12, while a greedy detector selects the
center rectangle for a total score of only 7.}
\label{bad-overlap}
\end{center}
\end{figure}

It should be noted that solving Equation~(\ref{eq:detector}) exactly is not computationally feasible. 
Thus, this algorithm does not always find the optimal solution to (\ref{eq:detector}).  An
example which leads to suboptimal results is shown in Figure~\ref{bad-overlap}.  
However, as we will see, this suboptimal behavior does not lead to difficulties.    
Moreover, in the next section, we give an optimization algorithm capable of 
finding an appropriate window scoring function for use with Algorithm~\ref{alg:detector}.

\begin{algorithm}[tb]
   \caption{Object Detection}
   \label{alg:detector}
\begin{algorithmic}[1]
   \REQUIRE image $x$, window scoring function $f$
   \STATE $\mathcal{D}$ := all rectangles $r \in \mathcal{R}$ such that $f(x,r) > 0$
   \STATE Sort $\mathcal{D}$ such that $\mathcal{D}_1 \ge \mathcal{D}_2 \ge \mathcal{D}_3 \ge ...$
   \STATE $y^* := \{\}$
   \FOR{$i=1$ {\bfseries to} $|\mathcal{D}|$}
     \IF{$\mathcal{D}_i$ does not overlap any rectangle in $y^*$}
        \STATE $y^* := y^* \cup \{\mathcal{D}_i\}$
     \ENDIF
   \ENDFOR
   \STATE {\bfseries Return:} $y^*$, The detected object positions.
\end{algorithmic}
\end{algorithm}

\section{Max-Margin Object Detection}

In this work, we consider only window scoring functions which are linear in their parameters.  In particular, we
use functions of the form
\begin{equation}
f(x,r) = \langle w, \phi(x,r) \rangle
\end{equation}
where $\phi$ extracts a feature vector from the sliding window location $r$ in image $x$, and $w$ is a 
parameter vector.  If we denote the sum of window scores for a set of rectangles, $y$, as $F(x,y)$, then Equation~(\ref{eq:detector})
becomes
\begin{equation}
y^* = \argmax_{y \in \mathcal{Y}} F(x,y) = \argmax_{y \in \mathcal{Y}}\sum_{r \in y} \langle w, \phi(x,r)\rangle.
\end{equation}

Then we seek a parameter vector $w$ which leads to the fewest possible detection mistakes.  That
is, given a randomly selected image and label pair $(x_i,y_i) \in \mathcal{X}\times\mathcal{Y}$,
we would like the score for the correct labeling of $x_i$ to be
larger than the scores for all the incorrect labelings.  Therefore,
\begin{equation}
F(x_i, y_i) > \max_{y \neq y_i} F(x_i, y)
\end{equation}
should be satisfied as often as possible.

\subsection{The Objective Function for Max-Margin Object Detection}

Our algorithm takes a set of images $\{x_1, x_2, ..., x_n\} \subset \mathcal{X}$ 
and associated labels $\{y_1, y_2, ..., y_n\} \subset \mathcal{Y}$ and
attempts to find a $w$ such that the detector makes the correct prediction on each training
sample.  We take a max-margin approach \cite{Joachims/etal/09b} and require that the label for each training
sample is correctly predicted with a large margin. This leads to the following 
convex optimization problem:
\begin{align}
\label{eq:hard-margin}
\min_{w} & \quad  \frac{1}{2}||w||^2  & \\
\nonumber\text{s.t.} &  \quad F(x_i,y_i) \geq \max_{y\in\mathcal{Y}} \left[ F(x_i,y) + \triangle(y,y_i)\right],        &\forall i
\end{align}

Where $\triangle(y,y_i)$ denotes the loss for predicting a labeling of $y$ when the true 
labeling is $y_i$.   In particular, we define the loss as 
\begin{align}
\triangle(y,y_i) = & L_{miss} \cdot (\text{\# of missed detections}) + \\
          \nonumber         & L_{fa} \cdot (\text{\# of false alarms})
\end{align}
where $L_{miss}$ and $L_{fa}$ control the relative importance of achieving  high
recall and high precision, respectively.

Equation~(\ref{eq:hard-margin}) is a hard-margin formulation of our learning problem.  
Since real world data is often noisy, not perfectly separable, or contains outliers, we 
extend this into the soft-margin setting.  In doing so, we arrive at the defining optimization 
for Max-Margin Object Detection (MMOD)
\begin{align}
\label{eq:soft-margin}
\min_{w,\xi}  \quad & \frac{1}{2}||w||^2 + \frac{C}{n}\sum_{i=1}^n \xi_i & \\
\nonumber\text{s.t.} \quad  & F(x_i,y_i) \geq \max_{y\in\mathcal{Y}} \left[ F(x_i,y) + \triangle(y,y_i)\right] - \xi_i, &\forall i \\
\nonumber  & \xi_i \geq 0, \quad \forall i
\end{align}
In this setting, C is analogous to the usual support vector machine parameter and controls the trade-off 
between trying to fit the training data or obtain a large margin.  

Insight into this formulation can be gained by noting that 
each $\xi_i$ is an upper bound on the loss incurred by training example $(x_i,y_i)$.  
This can be seen as follows (let $g(x) = \argmax_{y\in\mathcal{Y}}F(x,y)$)
\begin{align}
\label{xi1}\xi_i & \geq \max_{y\in\mathcal{Y}} \left[F(x_i,y) + \triangle(y,y_i) \right] - F(x_i,y_i) \\
\label{xi2}\xi_i & \geq \left[F(x_i, g(x_i)) + \triangle(g(x_i),y_i) \right] - F(x_i,y_i)  \\
\label{xi3}\xi_i & \geq \triangle(g(x_i),y_i)
\end{align}
In the step from (\ref{xi1}) to (\ref{xi2}) we replace the max over $\mathcal{Y}$ with a particular
element, $g(x_i)$.  Therefore, the inequality continues to hold.  In going from (\ref{xi2}) to (\ref{xi3})
we note that $F(x_i,g(x_i)) - F(x_i,y_i) \geq 0$ since $g(x_i)$ is by definition the element of $\mathcal{Y}$
which maximizes $F(x_i,\cdot)$.

Therefore, the MMOD objective function defined by Equation~(\ref{eq:soft-margin}) is a convex upper bound
on the average loss per training image
\begin{equation}
 \frac{C}{n}\sum_{i=1}^n \triangle(\argmax_{y\in\mathcal{Y}}F(x_i,y),y_i).
\end{equation}
This means that, for example, if $\xi_i$ from Equation~(\ref{eq:soft-margin}) is driven to zero then
the detector is guaranteed to produce the correct output from the corresponding training example.

This type of max-margin approach has been used successfully in a number of other
domains.  An example is the Hidden Markov SVM \cite{Altun03hiddenmarkov}, which gives
state-of-the-art results on sequence labeling tasks.  Other examples include
multiclass SVMs and methods for learning probabilistic context free grammars \cite{Joachims/etal/09b}.

\subsection{Solving the MMOD Optimization Problem }

We use the cutting plane method \cite{joachims_svm_struct, teo_bmrm} to solve the
Max-Margin Object Detection optimization problem defined by Equation~(\ref{eq:soft-margin}).  Note that 
MMOD is equivalent to the following unconstrained problem
\begin{equation}
\min_w J(w) = \frac{1}{2}||w||^2 + R_{emp}(w) 
\end{equation}
where $R_{emp}(w)$ is
\begin{equation}
 \frac{C}{n}\sum_{i=1}^n\max_{y\in\mathcal{Y}} \left[F(x_i,y) + \triangle(y,y_i) - F(x_i,y_i) \right].
\end{equation}

\begin{figure}[ht]
\begin{center}
\centerline{\includegraphics[width=\columnwidth]{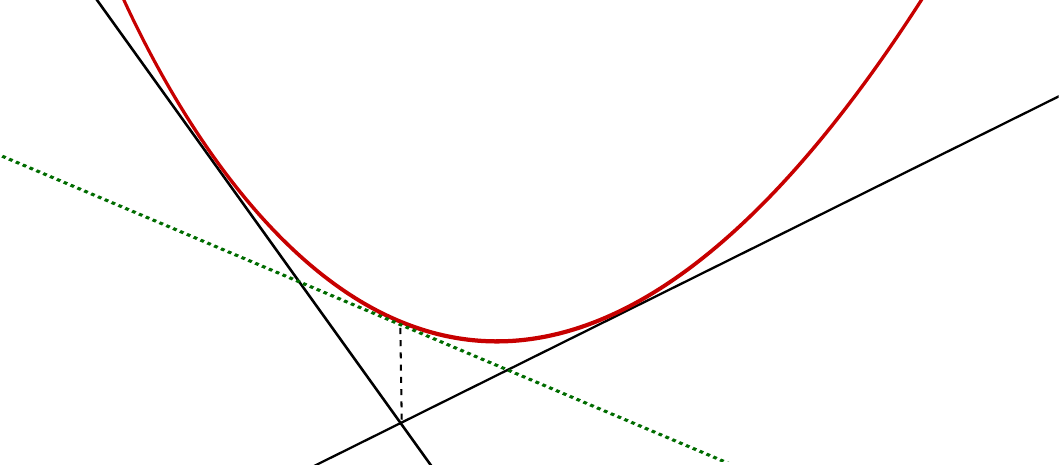}}
\caption{The red curve is lower bounded by its tangent planes.  Adding the tangent
plane depicted by the green line tightens the lower bound further.}
\label{fig:cpa}
\end{center}
\end{figure}

Further, note that $R_{emp}$ is a convex function of $w$ and therefore is lower bounded by any tangent plane.  
The cutting plane method exploits this to find the minimizer of $J$.  It does this by  building a 
progressively more accurate lower bounding approximation constructed from tangent planes.  Each step
of the algorithm finds a new $w$ minimizing this approximation.  Then it obtains the tangent plane to $R_{emp}$ at $w$,
and incorporates this new plane into the lower bounding function, tightening the approximation.  
A sketch of the procedure is shown in Figure~\ref{fig:cpa}.  

Let $\partial R_{emp}(w_t)$ denote the subgradient of $R_{emp}$ at a point $w_t$.
Then a tangent plane to $R_{emp}$ at $w_t$ is given by
\begin{equation}
\langle w, a \rangle + b
\end{equation}
where 
\begin{align}
&a \in \partial R_{emp}(w_t) \\
&b = R_{emp}(w_t) - \langle w_t,a\rangle.
\end{align}

Given these considerations, the lower bounding approximation we use is
\begin{equation}
\frac{1}{2}||w||^2 + R_{emp}(w) \geq \frac{1}{2}||w||^2 + \max_{(a,b)\in P}\left[\langle w,a\rangle + b\right]
\end{equation}
where $P$ is the set of lower bounding planes, i.e. the ``cutting planes''.

\begin{algorithm}[tb]
    \caption{MMOD Optimizer}
    \label{alg:cpa}
\begin{algorithmic}[1]
    \REQUIRE $\varepsilon \geq 0$
    \STATE $w_0 := 0$, $t := 0$, $P := \{\}$
    \REPEAT
        \STATE $t := t + 1$
        \STATE Compute plane tangent to $R_{emp}(w_{t-1})$, select $a_t \in\partial R_{emp}(w_{t-1})$
        and $b_t := R_{emp}(w_{t-1}) - \langle w_{t-1},a_t\rangle$
        \STATE $P_t := P_{t-1} \cup \{(a_t,b_t)\}$
        \STATE Let $K_t(w) = \frac{1}{2}||w||^2 + \max_{(a_i,b_i)\in P_t}[\langle w,a_i\rangle + b_i]$
        \STATE $w_t := \argmin_w K_t(w)$
    \UNTIL{$\frac{1}{2}||w_t||^2 + R_{emp}(w_t) - K_t(w_t) \leq \varepsilon$}
    \STATE {\bfseries Return:} $w_t$
\end{algorithmic}
\end{algorithm}

Pseudocode for this method is shown in Algorithm~\ref{alg:cpa}.  It 
executes until the gap between the true MMOD objective function and the
lower bound is less than $\varepsilon$.  This guarantees convergence
to the optimal $w^*$ to within $\varepsilon$.  That is, we will have
\begin{equation}
    |J(w^*) - J(w_t)| < \varepsilon
\end{equation}
upon termination of Algorithm~\ref{alg:cpa}.

\subsubsection{Solving the Quadratic Programming Subproblem}

A key step of Algorithm~\ref{alg:cpa} is solving the argmin on
step 7.  This subproblem can be written as a quadratic program and solved efficiently 
using standard methods.  Therefore, in this section we derive a simple quadratic
program solver for this problem.  We begin by writing step 7 as a quadratic program and obtain
\begin{align}
\min_{w,\xi} \quad &  \frac{1}{2}||w||^2 + \xi \\
\nonumber\text{s.t.} \quad  & \xi \geq \langle w,a_i\rangle + b_i, \quad \forall (a_i,b_i) \in P.
\end{align}

The set of variables being optimized, $w$, will typically have many more dimensions than the
number of constraints in the above problem.  Therefore, it is more efficient to solve the dual
problem.  To do this, note that the Lagrangian is 
\begin{equation}
L(w,\xi,\lambda) = \frac{1}{2} ||w||^2 + \xi - \sum_{i=1}^{|P|}\lambda_i(\xi - \langle w,a_i\rangle - b_i).
\end{equation}
and so the dual \cite{fletcher} of the quadratic program is
\begin{align}
\max_{w,\xi,\lambda} \quad & L(w,\xi,\lambda) \\
\nonumber\text{s.t.} \quad & \bigtriangledown_w L(w,\xi,\lambda) = 0, \\
\nonumber & \bigtriangledown_\xi L(w,\xi,\lambda) = 0, \\
\nonumber & \lambda_i \geq 0, \quad \forall i
\end{align}
After a little algebra, the dual reduces to the following quadratic program,
\begin{align}
\label{eq:dual-simple}
\max_{\lambda} \quad & \lambda^Tb - \frac{1}{2}\lambda^T Q \lambda  \\
\nonumber\text{s.t.} \quad  & \lambda_i \geq 0, \sum_{i=1}^{|P|} \lambda_i = 1
\end{align}
where $\lambda$ and $b$ are column vectors of the variables $\lambda_i$ and $b_i$ respectively
and $Q_{ij} = \langle a_i,a_j\rangle$.

\begin{algorithm}[tb]
    \caption{Quadratic Program Solver for Equation~(\ref{eq:dual-simple})}
    \label{alg:qp-solver}
\begin{algorithmic}[1]
    \REQUIRE $Q,b,\lambda,\varepsilon_{qp} \geq 0$
    \STATE $\tau = 10^{-10}$
    \REPEAT
        \STATE $\bigtriangledown := Q\lambda - b$
        \STATE $big := -\infty$
        \STATE $little := \infty$
        \STATE $l := 0$
        \STATE $b := 0$
        \FOR{$i=1$ {\bfseries to} $|\mathcal{\bigtriangledown}|$}
            \IF {$\bigtriangledown_i > big$ {\bfseries and} $\lambda_i > 0$ }
                \STATE $big := \bigtriangledown_i$
                \STATE $b := i$
            \ENDIF
            \IF {$\bigtriangledown_i < little$ }
                \STATE $little := \bigtriangledown_i$
                \STATE $l := i$
            \ENDIF
        \ENDFOR
        \STATE $gap := \lambda^T\bigtriangledown - little$ 
        \STATE $z := \lambda_b + \lambda_l$
        \STATE $x := \max(\tau, Q_{bb} + Q_{ll} - 2Q_{bl})$
        \STATE $\lambda_b := \lambda_b - (big-little)/x$
        \STATE $\lambda_l := \lambda_l + (big-little)/x$
        \IF {$\lambda_b < 0$}
            \STATE $\lambda_b := 0$
            \STATE $\lambda_l := z$
        \ENDIF
    \UNTIL{ $gap \leq \varepsilon_{qp}$}
    \STATE {\bfseries Return:} $\lambda$
\end{algorithmic}
\end{algorithm}

We use a simplified variant of Platt's sequential minimal optimization method to
solve the dual quadratic program of Equation~(\ref{eq:dual-simple}) \cite{platt.smo}.  Algorithm~\ref{alg:qp-solver} contains the pseudocode.  In each
iteration, the pair of Lagrange multipliers ($\lambda_b$, $\lambda_l$) which most violate the KKT conditions
are selected (lines 6-13).  Then the selected pair is jointly optimized (lines 15-21).
The solver terminates when the duality gap is less than a threshold.

Upon solving for the optimal $\lambda^*$, the $w_t$ needed by step 7 of 
Algorithm~\ref{alg:cpa} is given by
\begin{align}
w_t &= -\sum_{i=1}^{|P|} \lambda_i^* a_i.
\end{align}
The value of $\min_w K(w)$ needed for the test for convergence
can be conveniently computed as 
\begin{align}
 {\lambda^*}^Tb - \frac{1}{2}||w_t||^2.
\end{align}

Additionally, there are a number of non-essential but useful implementation tricks.  In particular, 
the starting $\lambda$ should be initialized using the $\lambda$ from the previous
iteration of the MMOD optimizer.   Also, cutting planes typically become inactive
after a small number of iterations and can be safely removed.  A cutting
plane is inactive if its associated Lagrange multiplier is 0.  Our implementation
removes a cutting plane if it has been inactive for 20 iterations.

\subsubsection{Computing $R_{emp}$ and its Subgradient }

\begin{algorithm}[t]
   \caption{Loss Augmented Detection}
   \label{alg:loss_detector}
\begin{algorithmic}[1]
   \REQUIRE image $x$, true object positions $y$, weight vector $w$, $L_{miss}$, $L_{fa}$ 
   \STATE $\mathcal{D}$ := all rectangles $r \in \mathcal{R}$ such that $\langle w, \phi(x,r)\rangle + L_{fa} > 0$
   \STATE Sort $\mathcal{D}$ such that $\mathcal{D}_1 \ge \mathcal{D}_2 \ge \mathcal{D}_3 \ge ...$
   \STATE $s_r := 0, h_r := false, \quad \forall r \in y$
   \FOR{$i=1$ {\bfseries to} $|\mathcal{D}|$}
     \IF{$\mathcal{D}_i$ does not overlap $\{\mathcal{D}_{i-1}, \mathcal{D}_{i-2}, ... \}$}
        \IF{$\mathcal{D}_i$ matches an element of $y$} 
            \STATE $r := $ best matching element of $y$
            \IF{$h_r = false$}
                \STATE $s_r := \langle w, \phi(x,\mathcal{D}_i)\rangle$
                \STATE $h_r := true$
            \ELSE
                \STATE $s_r := s_r + \langle w, \phi(x,\mathcal{D}_i)\rangle + L_{fa}$
            \ENDIF
        \ENDIF
     \ENDIF
   \ENDFOR
   \STATE $y^* := \{\}$
   \FOR{$i=1$ {\bfseries to} $|\mathcal{D}|$}
     \IF{$\mathcal{D}_i$ does not overlap $y^*$}
        \IF{$\mathcal{D}_i$ matches an element of $y$} 
            \STATE $r := $ best matching element of $y$
            \IF{$s_r > L_{miss}$}
            \STATE $y^* := y^* \cup \{\mathcal{D}_i\}$
            \ENDIF
        \ELSE
            \STATE $y^* := y^* \cup \{\mathcal{D}_i\}$
        \ENDIF
     \ENDIF
   \ENDFOR
   \STATE {\bfseries Return:} $y^*$, The detected object positions.
\end{algorithmic}
\end{algorithm}

The final component of our algorithm is a method for computing $R_{emp}$ and
an element of its subgradient.  Recall that $F(x,y)$ and $R_{emp}$ are
\begin{align}
 F(x,y) &= \sum_{r\in y} \langle w, \phi(x,r) \rangle \\
 R_{emp}(w) &= \frac{C}{n}\sum_{i=1}^n\max_{y\in\mathcal{Y}} \left[F(x_i,y) + \triangle(y,y_i) - F(x_i,y_i) \right].
\end{align}
Then an element of the subgradient of $R_{emp}$ is
\begin{equation}
\partial R_{emp}(w) = \frac{C}{n}\sum_{i=1}^n \left[ \sum_{r\in{y_i^*}}\phi(x_i,r)-\sum_{r\in{y_i}}\phi(x_i,r)  \right]
\end{equation}
where 
\begin{equation}
\label{eq:loss_detector}
y_i^* = \argmax_{y\in\mathcal{Y}} \left[\triangle(y,y_i) + \sum_{r\in y}\langle w, \phi(x_i,r) \rangle \right].
\end{equation}

Our method for computing $y_i^*$ is shown in Algorithm~\ref{alg:loss_detector}.  It is a
modification of the normal object detection procedure from Algorithm~\ref{alg:detector}
to solve Equation~(\ref{eq:loss_detector}) rather than (\ref{eq:detector}).  
Therefore, the task of Algorithm~\ref{alg:loss_detector} is to find the set of rectangles 
which jointly maximize the total detection score and loss.   

There are two cases to consider.  First, if a rectangle does not hit any truth rectangles, then
it contributes positively to the argmax in Equation~\ref{eq:loss_detector} whenever its score plus the loss per false alarm ($L_{fa}$)
is positive.  Second, if a rectangle hits a truth rectangle then we reason as
follows:  if we reject the first rectangle which matches a truth rectangle
then, since the rectangles are sorted in descending order of score, we will
reject all others which match it as well.  This outcome results in a single
value of $L_{miss}$.  Alternatively, if we accept the first rectangle which
matches a truth rectangle then we gain its detection score.  Additionally, we
may also obtain additional scores from subsequent duplicate detections, each of
which contributes the value of its window scoring function plus $L_{fa}$.
Therefore, Algorithm~\ref{alg:loss_detector} computes the total score for the
accept case and checks it against $L_{miss}.$  It then selects the result with
the largest value.  In the pseudocode, these scores are accumulated in the
$s_r$ variables.

This algorithm is greedy and thus may fail to find the optimal $y_i^*$ according to 
Equation~(\ref{eq:loss_detector}).   However, it is greedy in much the same way 
as the detection method of Algorithm~\ref{alg:detector}.  Moreover, since our goal from the
outset is to find a set of parameters which makes Algorithm~\ref{alg:detector} perform
well, we should use a training procedure which respects the properties of the
algorithm being optimized.  For example, if the correct output in the case of
Figure~\ref{bad-overlap} was to select the two boxes on the sides, then Algorithm~\ref{alg:detector}
would make a mistake while a method which was optimal would not.  Therefore,
it is important for the learning procedure to account for this and learn that in such
situations, if Algorithm~\ref{alg:detector} is to produce the correct output, the side rectangles need a larger
score than the middle rectangle.  
Ultimately, it is only necessary for Algorithm~\ref{alg:loss_detector} to give a value of $R_{emp}$ 
which upper bounds the average loss per training image.  In our experiments, we always observed this to be 
the case.

\section{Experimental Results}

To test the effectiveness of MMOD, we evaluate it on the TU Darmstadt cows \cite{tu_cows},
INRIA pedestrians \cite{hog}, and FDDB \cite{fddb} datasets.
When evaluating on the first two datasets, we use the same feature extraction
($\phi$) and parameter settings ($C$, $\varepsilon$, $\varepsilon_{qp}$,
$L_{fa}$, and $L_{miss}$), which are set as follows: $C = 25$, $\varepsilon = 0.15 C$, $L_{fa} = 1$, and $L_{miss} = 2$.
This value of $\varepsilon$ means the optimization runs until the potential improvement
in average loss per training example is less than $0.15$.  For the QP subproblem, we 
set $\varepsilon_{qp} = \min(0.01, 0.1(\frac{1}{2}||w_t||^2 + R_{emp}(w_t) - K_t(w_t)))$ to allow
the accuracy with which we solve the subproblem to vary as the overall optimization progresses.

For feature extraction, we use the popular spatial pyramid bag-of-visual-words model \cite{spatial_pyramid}.  
In our implementation, each window is divided into a 6x6 grid. Within  
each grid location we extract a 2,048 bin histogram of visual-words.    
The visual-word histograms are computed by extracting 36-dimensional HOG \cite{hog} descriptors from each pixel location, 
determining which histogram bin the feature is closest too, and adding 1 to that visual-word's bin count.  
Next, the visual-word histograms are concatenated to form the feature vector for the sliding 
window. Finally, we add a constant term which serves as a threshold for detection. Therefore, 
$\phi$ produces 73,729 dimensional feature vectors. 

The local HOG descriptors are 36 dimensional and are extracted from 10x10 grayscale pixel blocks
of four 5x5 pixel cells.  Each cell contains 9 unsigned orientation bins.  Bilinear 
interpolation is used for assigning votes to orientation bins but not for spatial
bins.  

To determine which visual-word a HOG descriptor corresponds to, many researchers
compute its distance to an exemplar for each bin and assign the vector to the 
nearest bin.  However, this is computationally expensive, so we use a fast approximate method  
to determine bin assignment. In particular, we use a random projection based locality
sensitive hash \cite{lsh}. 
This is accomplished using 11 random planes.  A HOG vector
is hashed by recording the bit pattern describing which side of each plane it falls on.  This
11-bit number then indicates the visual-word's bin.

Finally, the sliding window classification can be implemented efficiently using a set of integral images.
We also scan the sliding window over every location in an image pyramid which downsamples each 
layer by $4/5$.  To decide if two detections overlap for the purposes of non-max suppression we
use Equation~(\ref{eq:overlap}).  Similarly, we use Equation~(\ref{eq:overlap}) to determine 
if a detection hits a truth box.  Finally, all experiments were run on a single desktop workstation.

\subsection{TU Darmstadt Cows}


We performed 10-fold cross-validation on the TU Darmstadt cows \cite{tu_cows}
dataset and obtained perfect detection results with no false alarms.  The best
previous results on this dataset achieve an accuracy of 98.2\% at equal error rate \cite{Blaschko08}. 
The dataset contains 112 images, each containing a side-view of a cow.

In this test the sliding window was 174 pixels wide and 90 tall. 
Training on the entire cows dataset finishes in 49 iterations and takes 70 seconds.

\subsection{INRIA Pedestrians}

We also tested MMOD on the INRIA pedestrian dataset and followed the testing 
methodology used by Dalal and Triggs \cite{hog}.  
This dataset has 2,416 cropped images of people for training as well as 912 negative images.  
For testing it has 1,132 people images and 300 negative images.  

\begin{figure}[ht]
\begin{center}
\scalebox{0.75}{
\centerline{\includegraphics[width=\columnwidth]{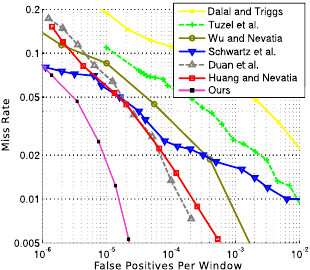}}}
\caption{The y axis measures the miss
rate on people images while the x axis shows FPPW obtained when scanning the detector over
negative images.  Our method improves both miss rate and false positives per window compared 
to previous methods on the INRIA dataset.}
\label{inria-results}
\end{center}
\end{figure}

The negative testing images have an average of 199,834 pixels per image.  We scan our detector 
over an image pyramid which downsamples at a rate of 4/5 and stop when the smallest pyramid
layer contains 17,000 pixels.  Therefore, MMOD scans approximately 930,000 windows per negative
image on the testing data.  

%

We use a sliding window 56 pixels wide and 120 tall.  The entire optimization takes  
116 minutes and runs for 220 iterations.
Our results are compared to
previous methods in Figure~\ref{inria-results}.  The detection tradeoff curve shows 
that our method achieves superior performance even though we use a basic bag-of-visual-word feature set
while more recent work has invested heavily in improved feature
representations.  Therefore, we attribute the increased improvement to our
training procedure.

\subsection{FDDB}

\begin{figure}[ht]
\begin{center}
\scalebox{1.00}{
\centerline{\includegraphics[scale=0.5]{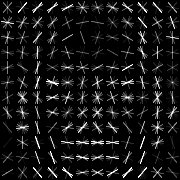}}}
\caption{The HOG filter learned via MMOD from the first fold of the FDDB dataset.  The filters
from other folds look nearly identical.}
\label{hog-filter}
\end{center}
\end{figure}

Finally, we evaluate our method on the Face Detection Data Set and Benchmark (FDDB) challenge.   
This challenging dataset contains images of human faces in multiple poses captured in indoor and outdoor settings.
To test MMOD, we used it to learn a basic HOG sliding window classifier. Therefore the feature extractor ($\phi$) takes in
a window and outputs a HOG vector describing the entire window as was done in Dalal and Triggs's seminal paper\cite{hog}.  
To illustrate the learned model, the HOG filter resulting from the first FDDB fold is visualized in Figure~\ref{hog-filter}.  

During learning, the parameters were set as follows: $C = 50$, $\varepsilon = 0.01 C$, $L_{fa} = 1$, and $L_{miss} = 1$.
We also upsampled each image by a factor of two so that smaller faces could be detected.  Since our HOG filter box is 80x80 pixels
in size this upsampling allows us to detect images that are larger than about 40x40 pixels in size.  Additionally, we mirrored the dataset, effectively
doubling the number of training images.  This leads to training sizes of about 5000 images per fold and our optimizer requires
approximately 25 minutes per fold.

\begin{figure}[ht]
\begin{center}
\scalebox{0.90}{
\centerline{\includegraphics[width=\columnwidth]{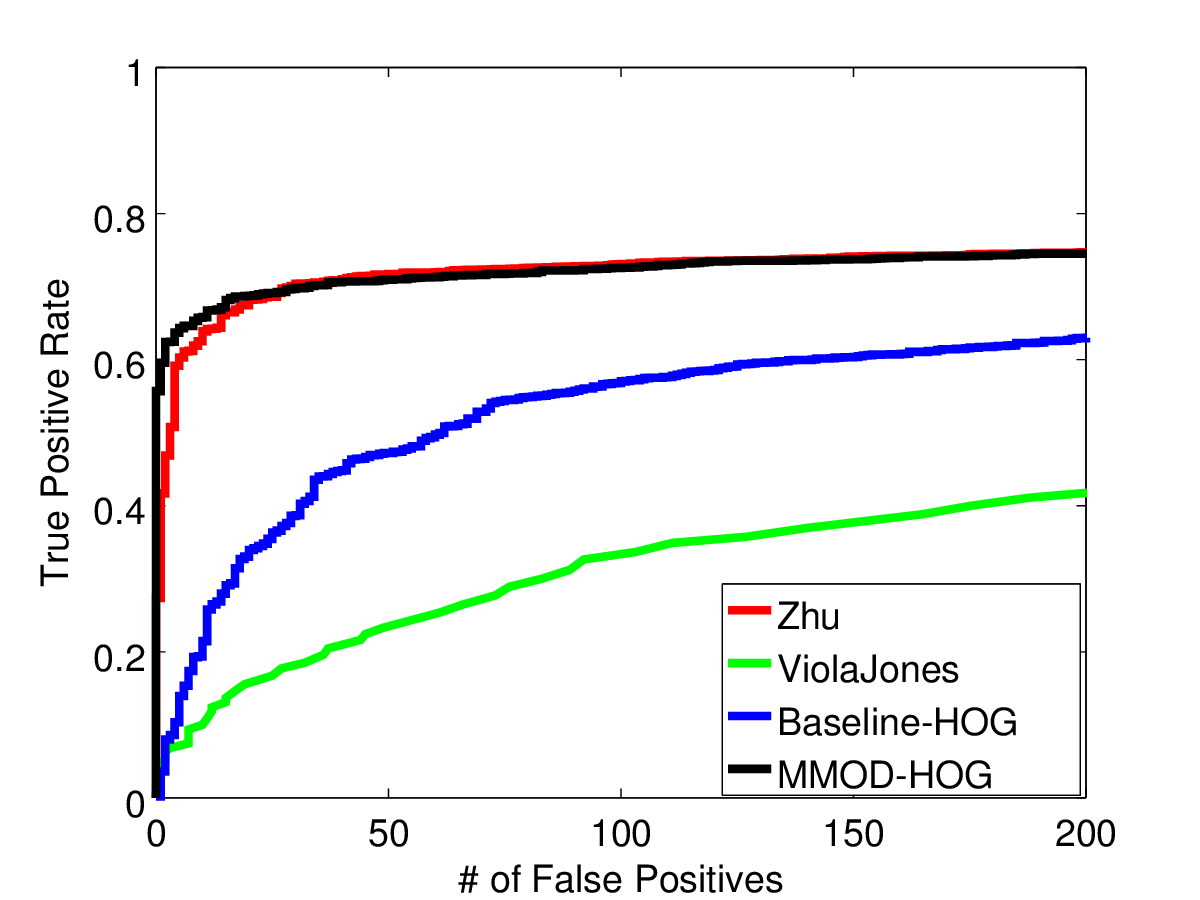}}}
\caption{
    A comparison between our HOG filter learned via MMOD and three other techniques, including another HOG
    filter method learned using traditional means.  The MMOD procedure results in a much more accurate HOG filter.
}
\label{fddb-results}
\end{center}
\end{figure}

\begin{figure*}[ht]
\begin{center}
\scalebox{0.175}{
\minibox{
\includegraphics[scale=1.99]{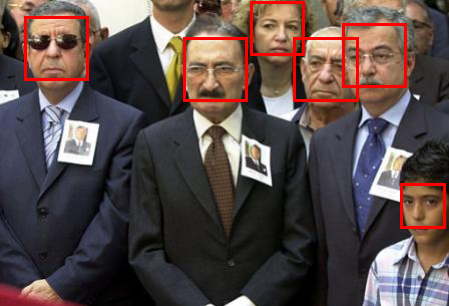}
\includegraphics[scale=1.07]{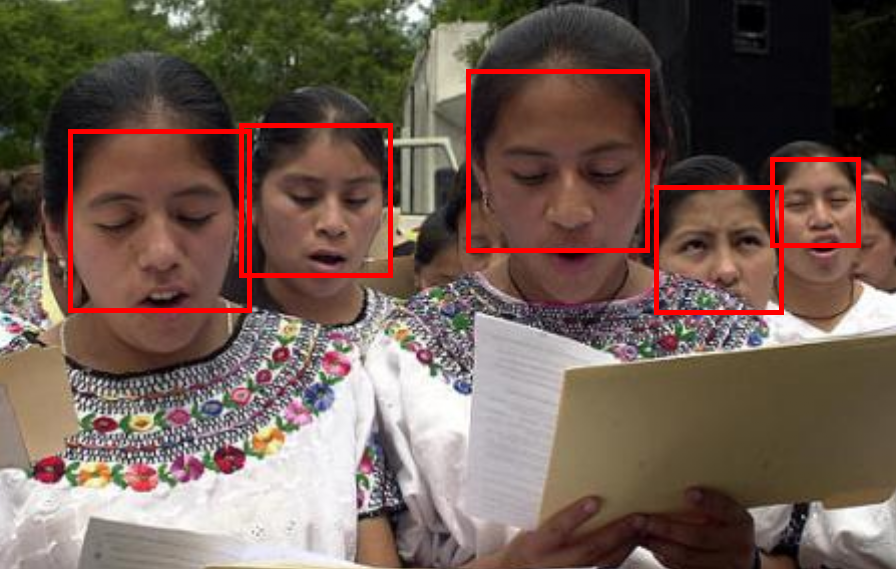}
\includegraphics[scale=0.80]{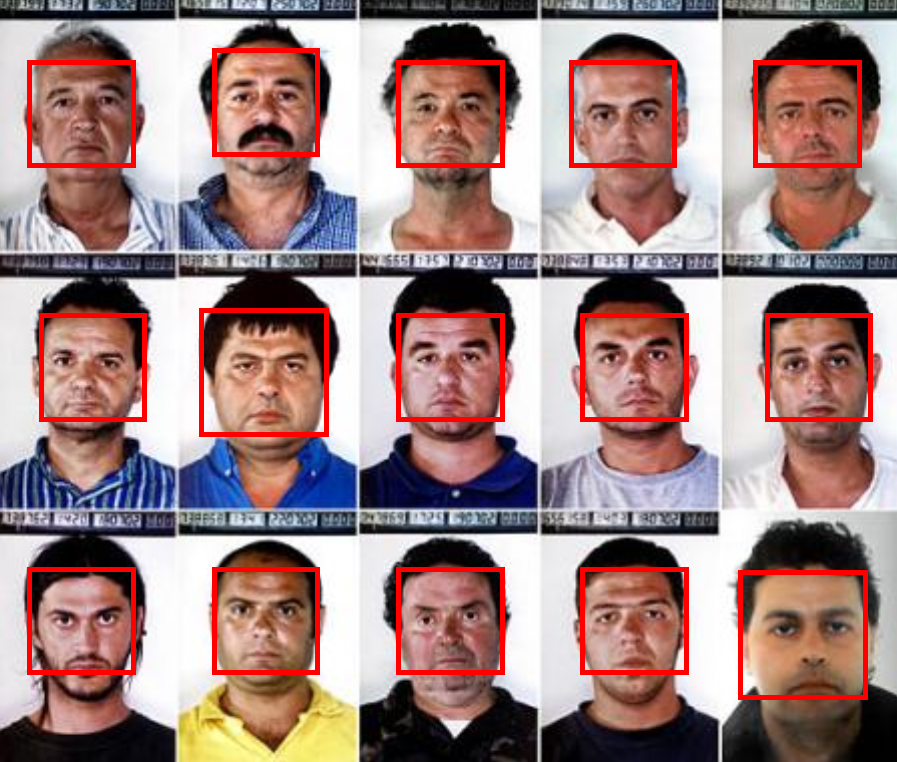}
}
}
\caption{Example images from the FDDB dataset.  The red boxes show the detections from HOG filters
    learned using MMOD.  The HOG filters were not trained on the images shown. }
\label{fddb-pics}
\end{center}
\end{figure*}

To perform detection, this single HOG filter is scanned over the image at each level of an image pyramid and any
windows which pass a threshold test are output after non-max suppression is performed.  A ROC curve that compares this learned HOG filter against other
methods is created by sweeping this threshold and can be seen in Figure~\ref{fddb-results}.  To create the ROC curve
we followed the FDDB evaluation protocol of performing 10 fold cross-validation and combining the results in a single
ROC curve using the provided FDDB evaluation software.
Example images with detection outputs are also shown in Figure~\ref{fddb-pics}. 

In Figure~\ref{fddb-results} we see that the HOG filter learned via MMOD substantially
outperforms a HOG filter learned with the typical linear SVM ``hard negative mining'' 
approach\cite{kostingerrobust} as well as the classic Viola Jones
method \cite{viola2001rapid}.  Moreover, our single HOG filter learned via MMOD
gives a slightly better accuracy than the complex deformable part model of Zhu
\cite{zhu2012face}.

\section{Conclusion}

We introduced a new method for learning to detect objects in images.  In
particular, our method leads to a convex optimization and we provided an
efficient algorithm for its solution.  We tested our
approach on three publicly available datasets, the INRIA person dataset, TU
Darmstadt cows, and FDDB using two feature representations.  On all
datasets, using MMOD to find the parameters of the detector lead to
substantial improvements.

Our results on FDDB are most striking as we showed that
a single rigid HOG filter can beat a state-of-the-art deformable part model
when the HOG filter is learned via MMOD.  We attribute our success to
the learning method's ability to make full use of the data.  In particular, on
FDDB, our method can efficiently make use of all 300 million sliding window
positions during training.  Moreover, MMOD optimizes the overall accuracy of
the entire detector, taking into account information which is typically ignored
when training a detector.  This includes windows which partially overlap target
windows as well as the non-maximum suppression strategy used in the final
detector.  

Our method currently uses a linear window scoring function.  Future
research will focus on extending this method to use more-complex
scoring functions, possibly by using kernels.  The work of Yu and Joachims
is a good starting point \cite{Yu/Joachims/08b}.  Additionally, while
our approach was introduced for 2D sliding window problems, it
may also be useful for 1D sliding window detection applications, such as
those appearing in the speech and natural language processing domains.
Finally, to encourage future research, we have made a 
careful and thoroughly documented implementation of our method available 
as part of the open source dlib\footnote{Software available at http://dlib.net/ml.html\#structural\_object\_detection\_trainer} machine learning toolbox \cite{dlib09}.


{\small
\bibliographystyle{ieee}
\bibliography{egbib}
}

\end{document}